\documentclass[10pt,twocolumn,letterpaper]{article}

\usepackage{subfigure}
\usepackage{cvpr}
\usepackage{times}
\usepackage{epsfig}
\usepackage{graphicx}
\usepackage{amsmath}
\usepackage{amssymb}
\usepackage{mathrsfs}
\usepackage{bm}
\usepackage{multirow}
\usepackage{array}

\usepackage[breaklinks=true,bookmarks=false]{hyperref}

\cvprfinalcopy % *** Uncomment this line for the final submission

\def\cvprPaperID{****} % *** Enter the CVPR Paper ID here

\ifcvprfinal\fi
\begin{document}

%%%%%%%%% TITLE
\title{Multi-Frame Quality Enhancement for Compressed Video}

\author{Ren Yang, Mai Xu\thanks{Mai Xu is the corresponding author of this paper.}, Zulin Wang and Tianyi Li\\
School of Electronic and Information Engineering, Beihang University, Beijing, China\\
{\tt\small \{yangren, maixu, wzulin, tianyili\}@buaa.edu.cn}
% For a paper whose authors are all at the same institution,
% omit the following lines up until the closing ``}''.
% Additional authors and addresses can be added with ``\and'',
% just like the second author.
% To save space, use either the email address or home page, not both
%\and
%Second Author\\
%Institution2\\
%First line of institution2 address\\
%{\tt\small secondauthor@i2.org}
}

\maketitle
%\thispagestyle{empty}

%%%%%%%%% ABSTRACT
\begin{abstract}
The past few years have witnessed great success in applying deep learning to enhance the quality of compressed image/video. The existing approaches mainly focus on enhancing the quality of a single frame, ignoring the similarity between consecutive frames. In this paper, we investigate that heavy quality fluctuation exists across compressed video frames, and thus low quality frames can be enhanced using the neighboring high quality frames, seen as Multi-Frame Quality Enhancement (MFQE). Accordingly, this paper proposes an MFQE approach for compressed video, as a first attempt in this direction. In our approach, we firstly develop a Support Vector Machine (SVM) based detector to locate Peak Quality Frames (PQFs) in compressed video. Then, a novel Multi-Frame Convolutional Neural Network (MF-CNN) is designed to enhance the quality of compressed video, in which the non-PQF and its nearest two PQFs are as the input. The MF-CNN compensates motion between the non-PQF and PQFs through the Motion Compensation subnet (MC-subnet). Subsequently, the Quality Enhancement subnet (QE-subnet) reduces compression artifacts of the non-PQF with the help of its nearest PQFs. Finally, the experiments validate the effectiveness and generality of our MFQE approach in advancing the state-of-the-art quality enhancement of compressed video. The code of our MFQE approach is available at \url{https://github.com/ryangBUAA/MFQE.git}.
\end{abstract}
\vspace{-2em}
%%%%%%%%% BODY TEXT
\section{Introduction}
\vspace{-.5em}
During the past decades, video has become significantly popular over the Internet. According to the Cisco Data Traffic Forecast \cite{Cisco}, video generates 60\% of Internet traffic in 2016, and this figure is predicted to reach 78\% by 2020. When transmitting video over the bandwidth-limited Internet, video compression has to be applied to significantly save the coding bit-rate \cite{Sikora2002The, wiegand2003overview,sullivan2012overview, li2017optimal}. However, the compressed video inevitably suffers from compression artifacts \cite{zeng2014characterizing, vlachos2000detection, li2015novel, yang2016subjective, yang2018saliency}, which may severely degrade the Quality of Experience (QoE). Therefore, it is necessary to study on quality enhancement for compressed video.

\begin{figure}[!t]
\centering
\vspace{-1em}
\includegraphics[width = 1\linewidth]{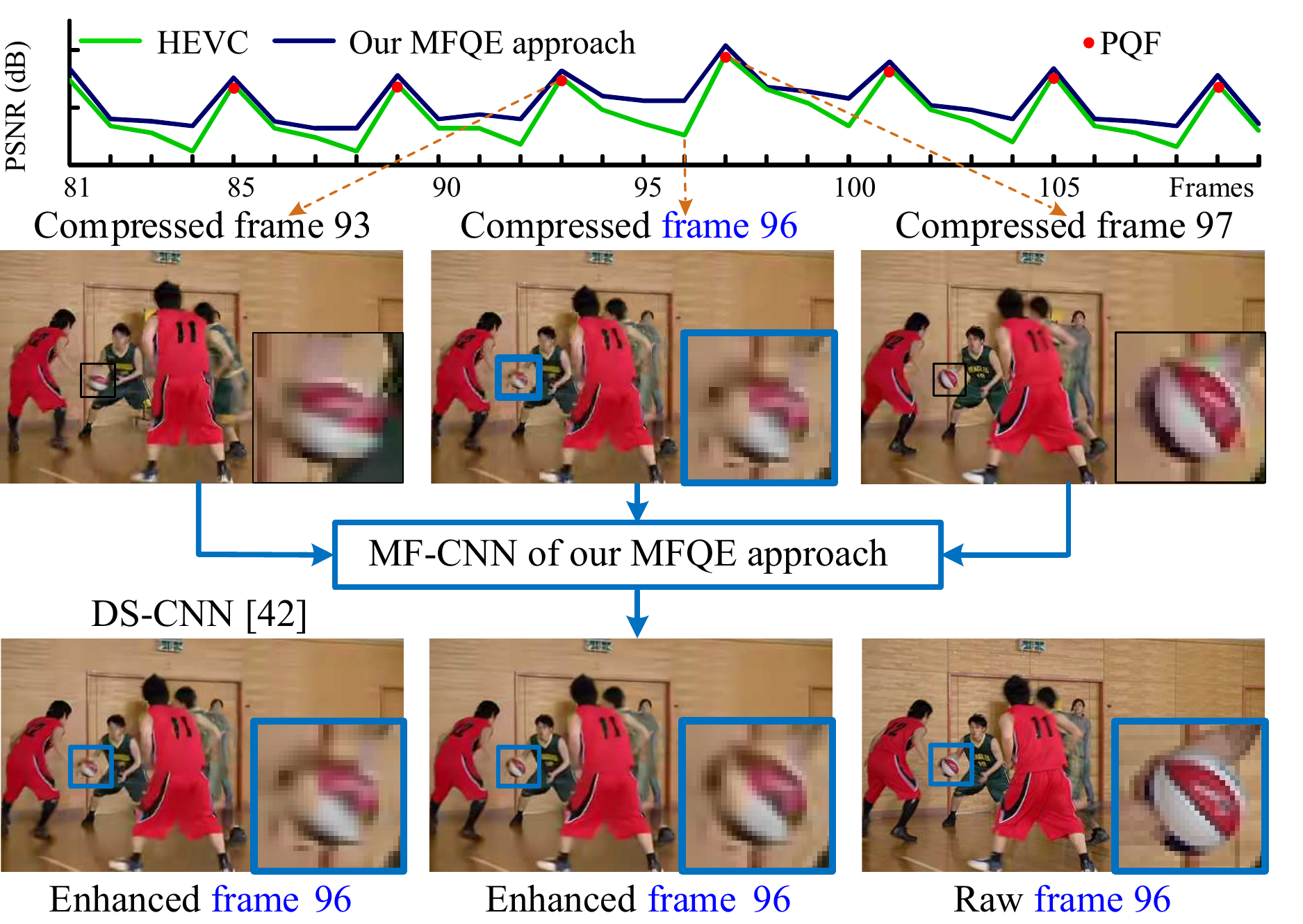}
\caption{\footnotesize{Example for quality fluctuation (top) and quality enhancement performance (bottom).}}\label{1}
\vspace{-2em}
\end{figure}

Recently, there has been increasing interest in enhancing the visual quality of compressed image/video \cite{liew2004blocking,foi2007pointwise,jancsary2012loss,wang2013adaptive,chang2014reducing,dong2015compression,wang2016d3,Guo2016Building,li2017efficient,li2017efficient,yang2017decoder}. For example, Dong \textit{et al.} \cite{dong2015compression} designed a four-layer Convolutional Neural Network (CNN) \cite{lecun1998gradient}, named AR-CNN, which considerably improves the quality of JPEG images. Later, Yang \textit{et al.} designed a Decoder-side Scalable CNN (DS-CNN) for video quality enhancement \cite{yang2017decoder, yang2017enhancing}.
%DS-CNN is composed of two subnets, which are designed to relieve intra- and inter-coding distortion, respectively.
However, when processing a single frame, all existing quality enhancement approaches do not take any advantage of information in the neighbouring frames, and thus their performance is largely limited. As Figure \ref{1} shows, the quality of compressed video dramatically fluctuates across frames. Therefore, it is possible to use the high quality frames (i.e., Peak Quality Frames, called PQFs\footnote{PQFs are defined as the frames whose quality is higher than their previous and subsequent frames.}) to enhance the quality of their neighboring low quality frames (non-PQFs). This can be seen as Multi-Frame Quality Enhancement (MFQE), similar to multi-frame super-resolution \cite{Kappeler2016Video, Caballero_2017_CVPR}.

This paper proposes an MFQE approach for compressed video. Specifically, we first investigate that there exists large quality fluctuation across frames, for video sequences compressed by almost all coding standards. Thus, it is necessary to find PQFs that can be used to enhance the quality of their adjacent non-PQFs. To this end, we train a Support Vector Machine (SVM) as a no-reference method to detect PQFs. Then, a novel Multi-Frame CNN (MF-CNN) architecture is proposed for quality enhancement, in which both the current frame and its adjacent PQFs are as the inputs. Our MF-CNN includes two components, i.e., Motion Compensation subnet (MC-subnet) and Quality Enhancement subnet (QE-subnet). The MC-subnet is developed to compensate the motion between the current non-PQF and its adjacent PQFs. The QE-subnet, with a spatio-temporal architecture, is designed to extract and merge the features of the current non-PQF and the compensated PQFs. Finally, the quality of the current non-PQF can be enhanced by QE-subnet that takes advantage of the high quality content in the adjacent PQFs. For example, as shown in Figure \ref{1}, the current non-PQF (frame 96) and the nearest PQFs (frames 93 and 97) are fed in to the MF-CNN of our MFQE approach. As a result, the low quality content (basketball) of the non-PQF (frame 96) can be enhanced upon the same content but with high quality in the neighboring PQFs (frames 93 and 97). Moreover, Figure \ref{1} shows that our MFQE approach also mitigates the quality fluctuation, because of the considerable quality improvement of non-PQFs.

The main contributions of this paper are: (1) We analyze the frame-level quality fluctuation of video sequences compressed by various video coding standards. (2) We propose a novel CNN-based MFQE approach, which can reduce the compression artifacts of non-PQFs by making use of the neighboring PQFs.
\vspace{-.5em}
\section{Related works}
\vspace{-.5em}
\textbf{Quality enhancement.} Recently, extensive works  \cite{liew2004blocking,foi2007pointwise,wang2013adaptive,jancsary2012loss,jung2012image,chang2014reducing,dong2015compression,Guo2016Building, wang2016d3,li2017efficient,Cavigelli2017CAS} have focused on enhancing the visual quality of compressed image. Specifically, Foi \textit{et al.} \cite{foi2007pointwise} applied pointwise Shape-Adaptive DCT (SA-DCT) to reduce the blocking and ringing effects caused by JPEG compression. Later, Jancsary \textit{et al.} \cite{jancsary2012loss} proposed reducing JPEG image blocking effects by adopting Regression Tree Fields (RTF). Moreover, sparse coding was utilized to remove the JPEG artifacts, such as \cite{jung2012image} and \cite{chang2014reducing}. Recently, deep learning has also been successfully applied to improve the visual quality of compressed image. Particularly, Dong \textit{et al.} \cite{dong2015compression} proposed a four-layer AR-CNN to reduce the JPEG artifacts of images. Afterwards, $\text{\textbf{D}}^3$ \cite{wang2016d3} and Deep Dual-domain Convolutional Network (DDCN) \cite{Guo2016Building} were proposed as advanced deep networks for the quality enhancement of JPEG image, utilizing the prior knowledge of JPEG compression. Later, DnCNN was proposed in \cite{Zhang2017Beyond} for several tasks of image restoration, including quality enhancement. Most recently, Li \textit{et al.} \cite{li2017efficient} proposed a 20-layer CNN, achieving the state-of-the-art quality enhancement performance for compressed image.

For the quality enhancement of compressed video, the Variable-filter-size Residue-learning CNN (VRCNN) \cite{dai2017convolutional} was proposed to replace the inloop filters for HEVC intra-coding. However, the CNN in \cite{dai2017convolutional} was designed as a component of the video encoder, so that it is not practical for already compressed video. Most recently, a Deep CNN-based Auto Decoder (DCAD), which contains 10 CNN layers, was proposed in \cite{Wang2017A} to reduce the distortion of compressed video. Moreover, Yang \textit{et al.} \cite{yang2017decoder} proposed the DS-CNN approach for video quality enhancement. In \cite{yang2017decoder}, DS-CNN-I and DS-CNN-B, as two subnetworks of DS-CNN, are used to reduce the artifacts of intra- and inter-coding, respectively. More importantly, the video encoder does not need to be modified when applying the DCAD \cite{Wang2017A} and DS-CNN \cite{yang2017decoder} approaches. Nevertheless, all above approaches can be seen as single-frame quality enhancement approaches, as they do not use any advantageous information available in the neighboring frames. Consequently, the video quality enhancement performance is severely limited.

\begin{figure*}[!t]
\centering
\vspace{-1em}
\subfigure{\includegraphics[width = 0.6\linewidth]{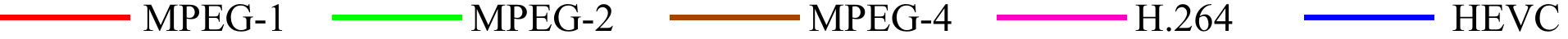}} \\
\vspace{-1em}
\subfigure{\includegraphics[width = 1\linewidth]{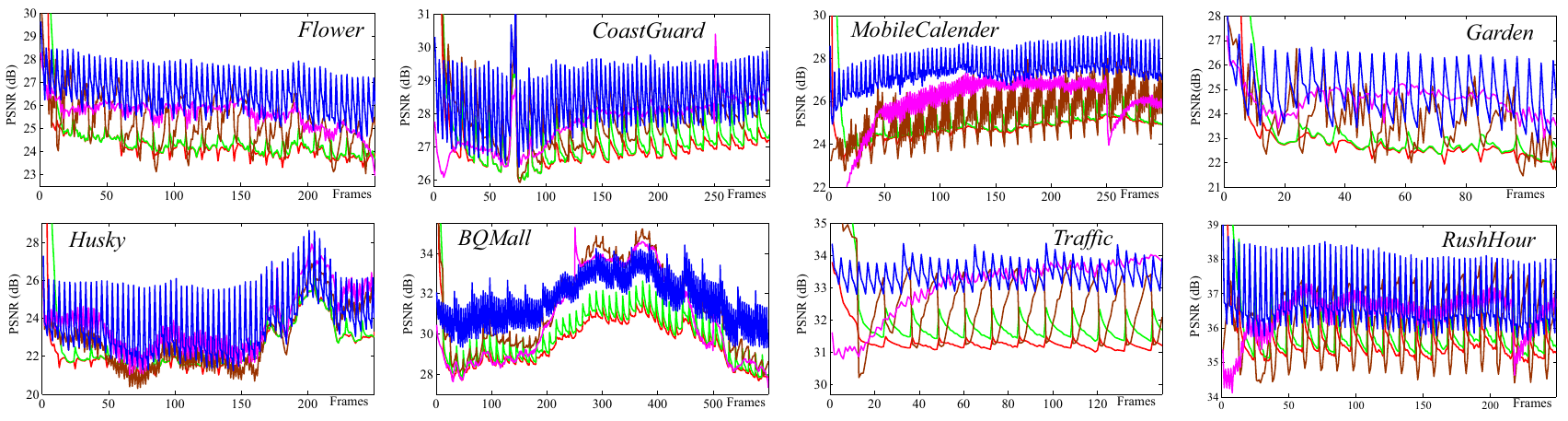}} \\
\caption{\small{PSNR curves of compressed video for various compression standards.}}\label{psnr}
\vspace{-1em}
\end{figure*}

\textbf{Multi-frame super-resolution.} To our best knowledge, there exists no MFQE work for compressed video. The closest area is multi-frame video super-resolution. In the early years, Brandi \textit{et al.} \cite{brandi2008super} and Song \textit{et al.} \cite{song2011video} proposed to enlarge video resolution by taking advantage of high resolution key-frames. Recently, many multi-frame super-resolution approaches have employed deep neural networks. For example, Huang \textit{et al.} \cite{Huang2015Bidirectional} developed a Bidirectional Recurrent Convolutional Network (BRCN), which improves the super-resolution performance over traditional single-frame approaches. In 2016, Kappeler \textit{et al.} proposed a Video Super-Resolution network (VSRnet) \cite{Kappeler2016Video}, in which the neighboring frames are warped according to the estimated motion, and both the current and warped neighboring frames are fed into a super-resolution CNN to enlarge the resolution of the current frame. Later, Li \textit{et al.} \cite{Li2017Video} proposed replacing VSRnet by a deeper network with residual learning strategy. Besides, other deep learning approaches of video super-resolution were proposed in \cite{Makansi2017End, Caballero_2017_CVPR}.

%In \cite{Makansi2017End}, the CNN-based FlowNet \cite{Dosovitskiy2015FlowNet,Ilg_2017_CVPR} is applied to estimate the motion across frames using the detected optical flow. Then, the motion estimation and super-resolution networks are trained jointly. In 2017, Caballero \textit{et al.} \cite{Caballero_2017_CVPR} designed a spatial transformer motion compensation network to detect the optical flow and warp neighboring frames. The current and warped neighboring frames are then input to the Efficient Sub-Pixel Convolution Network (ESPCN) \cite{Shi2016Real} for super-resolution.

The aforementioned multi-frame super-resolution approaches are motivated by the fact that different observations of the same objects or scenes are probably available across frames of a video. As a result, the neighboring frames may contain the content missed when down-sampling the current frame. Similarly, for compressed video, the low quality frames can be enhanced by taking advantage of their adjacent higher quality frames, because heavy quality fluctuation exists across compressed frames. Consequently, the quality of compressed video may be effectively improved by leveraging the multi-frame information. To the best of our knowledge, our MFQE approach proposed in this paper is the first attempt in this direction.
\vspace{-.5em}
\section{Quality fluctuation of compressed video}\label{quality}
\vspace{-.5em}
In this section, we analyze the quality fluctuation of compressed video alongside the frames. First, we establish a database including 70 uncompressed video sequences, selected from the datasets of Xiph.org \cite{Xiph} and JCT-VC \cite{bossen2011common}. We compress these sequences using various video coding standards, including MPEG-1 \cite{Gall1992The}, MPEG-2 \cite{Schafer1995Digital}, MPEG-4 \cite{Sikora2002The}, H.264/AVC \cite{wiegand2003overview} and HEVC \cite{sullivan2012overview}. The quality of each compressed frame is evaluated in terms of Peak Signal-to-Noise Ratio (PSNR).

Figure \ref{psnr} shows the PSNR curves of 8 video sequences, which are compressed by different standards. It can be seen that the compression quality obviously fluctuates along with the frames. We further measure the STandard Deviation (STD) of frame-level quality for each compressed video. As shown in Table \ref{tab:std}, the STD values of all five standards are above 1.00 dB, which are averaged over the 70 compressed sequences. The maximal STD among the 70 sequences reaches 3.97 dB, 4.00 dB, 3.84 dB, 5.67 dB and 3.34 dB for MPEG-1, MPEG-2, MPEG-4, H.264 and HEVC, respectively. This reflects the remarkable fluctuation of frame-level quality after video compression.

\begin{figure*}[!t]
\centering
%\subfigure{\includegraphics[width = 0.85\linewidth]{a.pdf}}
\subfigure{\includegraphics[width = 0.8\linewidth]{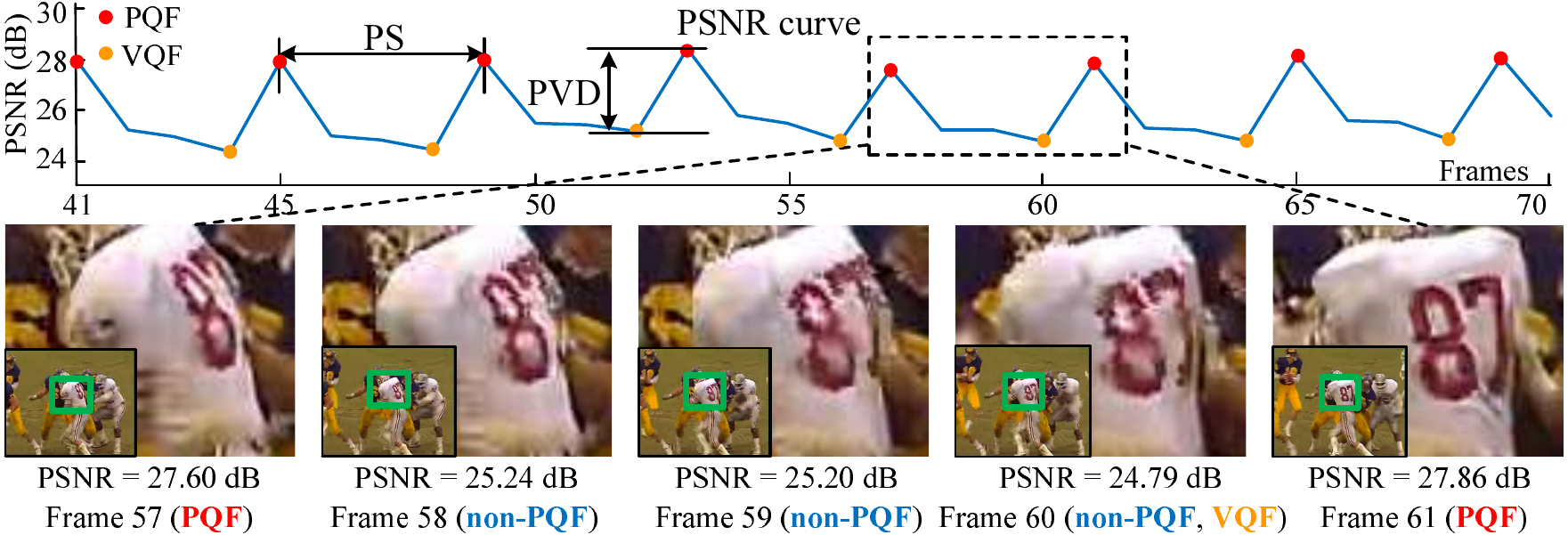}}
\caption{\small{Example of the frame-level PSNR and subjective quality for the HEVC compressed video sequence \textit{Football}.}}\label{squality}
\vspace{-1.5em}
\end{figure*}

Moreover, Figure \ref{squality} shows an example of the frame-level PSNR and subjective quality for one sequence, compressed by the latest HEVC standard. It can be observed from Figure \ref{squality} that there exists frequently alternate PQFs and Valley Quality Frames (VQFs). Here, PQF is defined as the frame whose quality is higher than its previous and subsequent frames. In contrast, VQF indicates the frame with lower quality than its previous and subsequent frames. As shown in this figure, the PSNR of non-PQFs (frames 58-60), especially the VQF (frame 60), is obviously lower than that of the nearest PQFs (frames 57 and 61). Moreover, non-PQFs (frames 58-60) also have much lower subjective quality than the nearest PQFs (frames 57 and 61), e.g., in the region of number ``87''. Additionally, the content of frames 57-61 is very similar. Hence, the visual quality of non-PQFs can be improved by using the content in the nearest PQFs.

%As shown in this figure, the PSNR of frame 60 (VQF) is 3.07 dB lower than that of the adjacent PQF, i.e., frame 61. Besides, the quality of other non-PQFs (frames 58 and 59) are also obviously lower than that of the nearest PQFs (frames 57 and 61). Figure \ref{squality}-(b) shows the subjective quality of PQFs and non-PQFs. As this figure shows, non-PQFs (frames 58-60) have much lower quality than their nearest PQFs (frames 57 and 61), especially in the region of number ``87''. Additionally, the content of frames 57-61 is very similar. Hence, the visual quality of non-PQFs can be improved by using the content in the nearest PQFs.

To further analyze the peaks and valleys of frame-level quality, we measure the Peak-Valley Difference (PVD) and Peak Separation (PS) for the PSNR curves of each compressed video sequence. As seen in Figure \ref{squality}-(a), PVD is denoted as the PSNR difference between the PQF and its nearest VQF, and PS indicates the number of frames between two PQFs. The averaged PVD and PS values of the 70 compressed video sequences are shown in Table \ref{tab:std} for each video coding standard. It can be seen that the averaged PVD values are higher than 1.00 dB in most cases, and the latest HEVC standard has the highest value of 1.50 dB. This verifies the large quality difference between PQFs and VQFs. Additionally, the PS values are approximately or less than 5 frames for each coding standard. In particular, the PS values are less than 3 frames for the H.264 and HEVC standards. Such a short distance between two PQFs indicates that the content of frames between the adjacent PQFs may be highly similar. Therefore, the PQFs probably contain some useful content which is distorted in their neighboring non-PQFs. Motivated by this, our MFQE approach is proposed to enhance the quality of non-PQFs through the advantageous information of the nearest PQFs.
\begin{table}[!t]
\vspace{-.5em}
  \centering
  \footnotesize
  \caption{\small{Averaged STD, PVD and PS values among our database.}}
    \begin{tabular}{|c|c|c|c|c|c|}
    \hline
          & MPEG-1 & MPEG-2 & MPEG-4 & H.264 & HEVC \\
    \hline
    STD (dB) & 1.8336  & 1.8347  & 1.7823  & 1.6370  & 1.0566  \\
    \hline
    PVD (dB) & 1.1072  & 1.1041  & 1.0420  & 0.5275 & 1.5089 \\
    \hline
    PS (frame) & 5.3438 & 5.3630  & 5.3819  &  1.6365  & 2.6600  \\
    \hline
    \end{tabular}%
  \label{tab:std}%
  \vspace{-2em}
\end{table}%
\section{The proposed MF-CNN approach}

\subsection{Framework}\label{frame}

\begin{figure*}[!t]
\centering
\includegraphics[width = .85\linewidth]{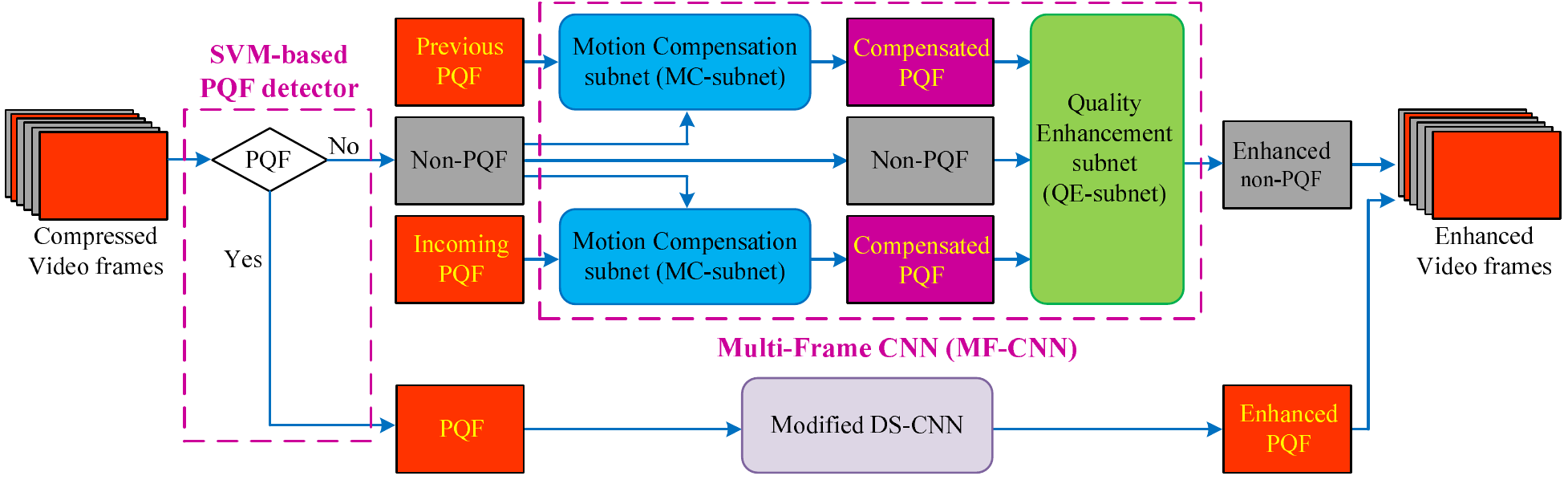}
\caption{\small{Framework of the proposed MFQE approach.}}\label{framework}
\vspace{-1.5em}
\end{figure*}

Figure \ref{framework} shows the framework of our MFQE approach. In the MFQE approach, we first detect the PQFs that are used for quality enhancement of other non-PQFs. In practical application, the raw sequences are not available in video quality enhancement, and thus the PQFs and non-PQFs cannot be distinguished through comparison with the raw sequences. Therefore, we develop a no-reference PQF detector in our MFQE approach, which is detailed in Section \ref{peakdetec}.

The quality of detected PQFs can be enhanced by DS-CNN \cite{yang2017decoder}, which is a single-frame approach for video quality enhancement. It is because the adjacent frames of a PQF are with lower quality and cannot benefit the quality enhancement of this PQF. Here, we modify the DS-CNN via replacing the Rectified Linear Units (ReLU) by Parametric ReLU (PReLU) to avoid zero gradients \cite{He2016Delving}, and we also apply residual learning \cite{he2016deep} to improve the quality enhancement performance.

For non-PQFs, the MF-CNN is proposed to enhance the quality that takes advantage of the nearest PQFs (i.e., both previous and subsequent PQFs). The MF-CNN architecture is composed of the MC-subnet and the QE-subnet. The MC-subnet is developed to compensate the temporal motion across the neighboring frames. To be specific, the MC-subnet firstly predicts the temporal motion between the current non-PQF and its nearest PQFs. Then, the two nearest PQFs are warped with the spatial transformer according to the estimated motion. As such, the temporal motion between the non-PQF and PQFs can be compensated. The MC-subnet is to be introduced in Section \ref{mc}.

Finally, the QE-subnet, which has a spatio-temporal architecture, is proposed for quality enhancement, as introduced in Section \ref{cnn}. In the QE-subnet, both the current non-PQF and the compensated PQFs are as the inputs, and then the quality of the current non-PQF can be enhanced under the help of the adjacent compensated PQFs. Note that, in the proposed MF-CNN, the MC-subnet and QE-subnet are trained jointly in an end-to-end manner.

\subsection{SVM-based PQF detector}\label{peakdetec}

In our MFQE approach, an SVM classifier is trained to achieve no-reference PQF detection. Recall that PQF is the frame with higher quality than the adjacent frames. Thus both the features of the current and four neighboring frames are used to detect PQFs. In our approach, the PQF detector follows the no-reference quality assessment method \cite{Mittal2012No} to extract 36 spatial features from the current frame, each of which is one-dimensional. Beyond, such kinds of spatial features are also extracted from two previous frames and two incoming frames. Consequently, 180 one-dimensional features are obtained to predict whether a frame is a PQF or non-PQF, based on the SVM classifier.

In our SVM classifier,  $l_n \in \{0, 1\}$ denotes the output class label indicating whether the $n$-th frame is a PQF (positive sample with $l_n=1$) or non-PQF (negative sample with $l_n=0$). We use the LIBSVM library \cite{CC01a} to train the SVM classifier, in which the probability of $l_n=1$ can be obtained for each frame and denoted as $p_n$. In our SVM classifier, the Radial Basis Function (RBF) is used as the kernel.

Finally, $\{l_n,p_n\}_{n=1}^N$ can be obtained from the SVM classifier, in which $N$ is the total number of frames in the video sequence. In our PQF detector, we further refine the results of the SVM classifier according to the prior knowledge of PQF. Specifically, the following two strategies are developed to refine the labels $\{l_n\}_{n=1}^N$ of the PQF detector.

(1) According to the definition of PQF, it is impossible that the PQFs consecutively appear. Hence, if the following case exists
\begin{eqnarray}\label{c1}
\vspace{-1em}
\{l_{n+i}\}_{i=0}^j = 1\ \ \ \text{and}\ \ \ l_{n-1}=l_{n+j+1}=0,\ j\geq1,
\vspace{-1em}
\end{eqnarray}
we set
\begin{eqnarray}\label{p1}
\vspace{-1em}
l_{n+i} = 0, \text{where}\ \ i \not= \mathop{\arg \max}_{0\leq k\leq j}(p_{n+k})
\vspace{-1em}
\end{eqnarray}
in our PQF detector.

(2) According to the analysis of Section \ref{quality}, PQFs frequently appear within a limited separation. For example, the average value of PS is 2.66 frames for HEVC compressed sequences. Here, we assume that $D$ is the maximal separation between two PQFs. Given this assumption, if the results of $\{l_n\}_{n=1}^N$ yields more than $D$ consecutive zeros (non-PQFs):
\begin{eqnarray}\label{c2}
\vspace{-1em}
\{l_{n+i}\}_{i=0}^d = 0\ \ \ \text{and}\ \ \ l_{n-1}=l_{n+d+1}=1,\ d > D,
\vspace{-1em}
\end{eqnarray}
one of frames need to be selected as PQF, and thus we set
\begin{eqnarray}\label{p2}
\vspace{-1em}
l_{n+i} = 1, \text{where}\ \ i = \mathop{\arg\max}_{0 < k < d}(p_{n+k}).
\vspace{-1em}
\end{eqnarray}
After refining $\{l_n\}_{n=1}^N$ as discussed above, our PQF detector can locate PQFs and non-PQFs in the compressed video.

\begin{figure}[!t]
\vspace{-.5em}
\centering
\includegraphics[width = 1\linewidth]{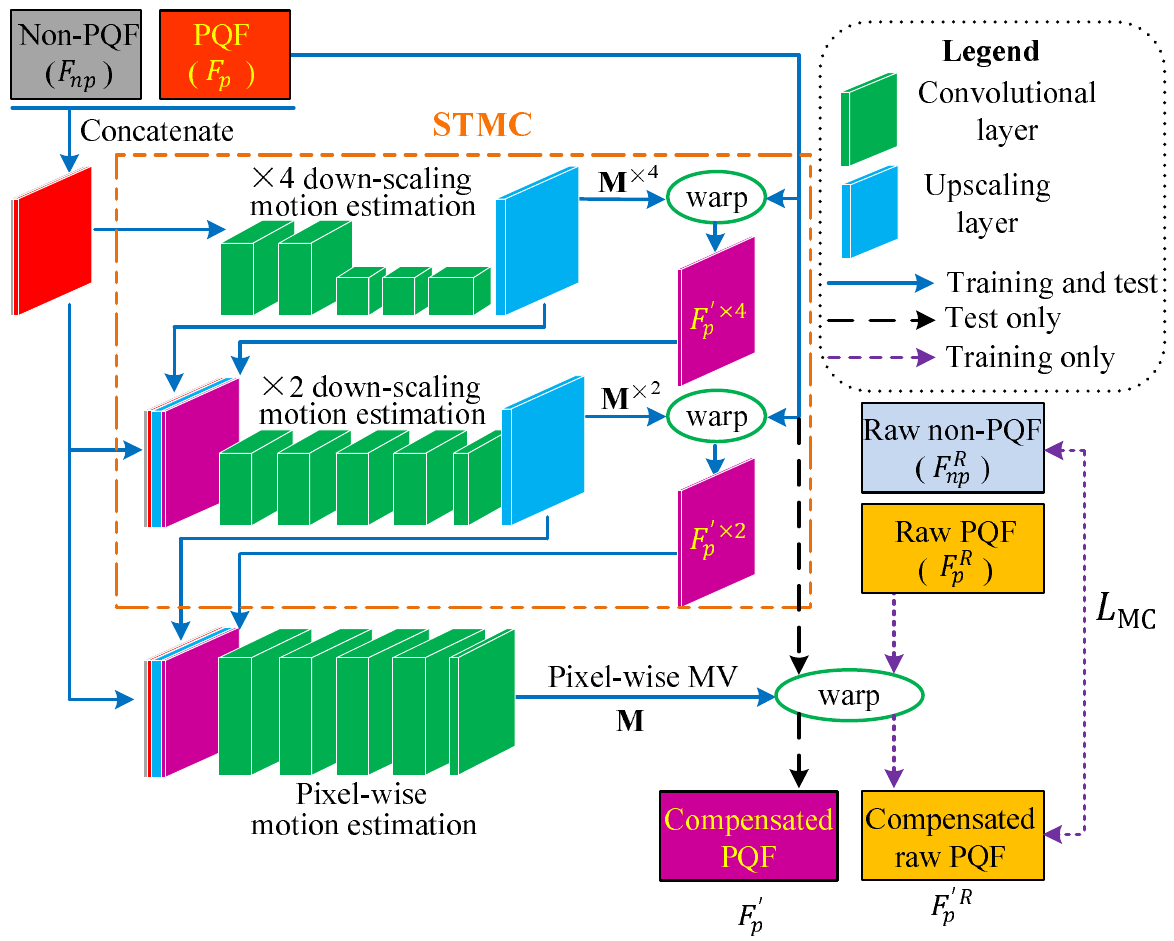}
\vspace{-1.5em}
\caption{\small{Architecture of our MC-subnet.}}\label{mcn}
\end{figure}

\subsection{MC-subnet}\label{mc}

After PQFs are detected, the quality of non-PQFs can be enhanced by taking advantage of the neighboring PQFs. However, the temporal motion exists between non-PQFs and PQFs. Hence, we develop the MC-subnet to compensate the temporal motion across frames. In the following, the architecture and training strategy of our MC-subnet are introduced in detail.

% Table generated by Excel2LaTeX from sheet 'Sheet1'
\begin{table}[!t]
  \centering
  \footnotesize
  \caption{\small{Convolutional layers for pixel-wise motion estimation.}}
    \begin{tabular}{|c|c|c|c|c|c|}
    \hline
    Layers &
      Conv 1 &
      Conv 2 &
      Conv 3 &
      Conv 4 &
      Conv 5
       \\
    \hline
    Filter size &
      $3\times3$ &
      $3\times3$ &
      $3\times3$ &
      $3\times3$ &
      $3\times3$
       \\
    \hline
    Filter number &
      24 &
      24 &
      24 &
      24 &
      2
       \\
    \hline
    Stride & 1 & 1& 1& 1& 1

       \\
    \hline
    Function &
      {PReLU} &
      {PReLU} &
      {PReLU} &
      {PReLU} &
      {Tanh}
       \\

    \hline
    \end{tabular}%
  \label{tab:conf1}%
  \vspace{-1.5em}
\end{table}

\textbf{Architecture.} In \cite{Caballero_2017_CVPR}, Caballero \textit{et al.} proposed the Spatial Transformer Motion Compensation (STMC) method for multi-frame super-resolution. As shown in Figure \ref{mcn}, the STMC method adopts the convolutional layers to estimate the $\times4$ and $\times2$ down-scaling Motion Vector (MV) maps, denoted as $\mathbf{M}^{\times4}$ and $\mathbf{M}^{\times2}$. In $\mathbf{M}^{\times4}$ and $\mathbf{M}^{\times2}$, the down-scaling is achieved by adopting some convolutional layers with the stride of 2. For details of these convolutional layers, refer to \cite{Caballero_2017_CVPR}.

The down-scaling motion estimation is effective to handle large scale motion. However, because of down-scaling, the accuracy of MV estimation is reduced. Therefore, in addition to STMC, we further develop some additional convolutional layers for pixel-wise motion estimation in our MC-subnet, which does not contain any down-scaling process. The convolutional layers of pixel-wise motion estimation are described in Table \ref{tab:conf1}. As Figure \ref{mcn} shows, the output of STMC includes the $\times2$ down-scaling MV map $\mathbf{M}^{\times2}$ and the corresponding compensated PQF $F'^{\times2}_{p}$. They are concatenated with the original PQF and non-PQF, as the input to the convolutional layers of the pixel-wise motion estimation. Then, the pixel-wise MV map can be generated, which is denoted as $\mathbf{M}$. Note that the MV map $\mathbf{M}$ contains two channels, i.e., horizonal MV map $\mathbf{M}_x$ and vertical MV map $\mathbf{M}_y$. Here, $x$ and $y$ are the horizonal and vertical index of each pixel. Given $\mathbf{M}_x$ and $\mathbf{M}_y$, the PQF is warped to compensate the temporal motion. Let the compressed PQF and non-PQF be $F_p$ and $F_{np}$, respectively. The compensated PQF $F'_p$ can be expressed as
\begin{eqnarray}
\vspace{-.5em}
F'_p(x,y) = \mathcal{I}\{F_{p}(x+\mathbf{M}_x(x,y),y+\mathbf{M}_y(x,y))\},
\vspace{-.5em}
\end{eqnarray}
where $\mathcal{I}\{\cdot\}$ means the bilinear interpolation. The reason for the interpolation is that $\mathbf{M}_x(x,y)$ and $\mathbf{M}_y(x,y)$ may be non-integer values.

\textbf{Training strategy.} Since it is hard to obtain the ground truth of MV, the parameters of the convolutional layers for motion estimation cannot be trained directly. The super-resolution work \cite{Caballero_2017_CVPR} trains the parameters by minimizing the MSE between the compensated adjacent frame and the current frame. However, in our MC-subnet, both the input $F_p$ and $F_{np}$ are compressed frames with quality distortion. Hence, when minimizing the MSE between $F'_p$ and the $F_{np}$, the MC-subnet learns to estimate the distorted MV, resulting in inaccurate motion estimation. Therefore, the MC-subnet is trained under the supervision of the raw frames. That is, we warp the raw frame of the PQF (denoted as $F^R_p$) using the MV map output from the convolutional layers of motion estimation, and minimize the MSE between the compensated raw PQF (denoted as $F'^{R}_{p}$) and the raw non-PQF (denoted as $F^R_{np}$). The loss function of the MC-subnet can be written by
\begin{eqnarray}
\vspace{-1em}
L_{\text{MC}}(\theta_{mc})=||F'^{R}_p(\theta_{mc}) - F^R_{np}||_2^2,
\vspace{-1em}
\end{eqnarray}
where $\theta_{mc}$ represents the trainable parameters of our MC-subnet. Note that the raw frames $F^R_p$ and $F^{R}_{np}$ are not required when compensating motion in the test.

\begin{figure}[!t]
\centering
\includegraphics[width = 1\linewidth]{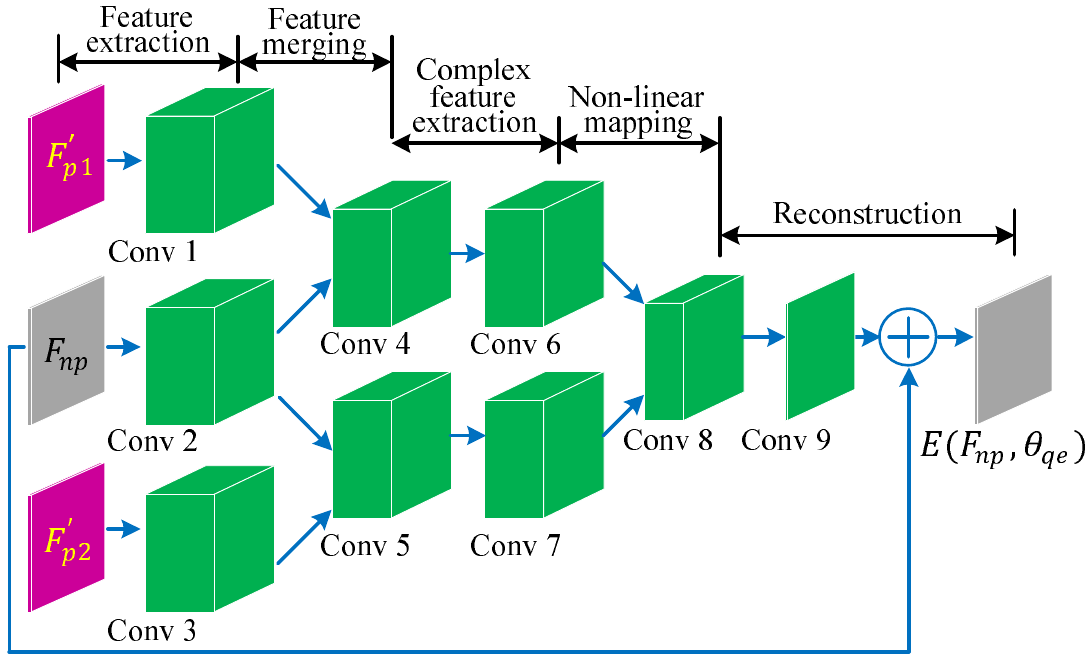}
\caption{\small{Architecture of our QE-subnet.}}\label{qen}
 \vspace{-1em}
\end{figure}

\begin{table}[!t]
  \centering
  \scriptsize
  \caption{\small{Convolutional layers of our QE-subnet.}}
    \begin{tabular}{|c|c|c|c|c|c|}
    \hline
    Layers &
      Conv 1/2/3 &
      Conv 4/5 &
      Conv 6/7 &
      Conv 8 &
      Conv 9
       \\
    \hline
    Filter size &
      $9\times9$ &
      $7\times7$ &
      $3\times3$ &
      $1\times1$ &
      $5\times5$
       \\
    \hline
    Filter number &
      128 &
      64 &
      64 &
      32 &
      1
       \\
    \hline
    Stride & 1 & 1& 1& 1& 1

       \\
    \hline
    Function &
      {PReLU} &
      {PReLU} &
      {PReLU} &
      {PReLU} &
      {---}
       \\
    \hline
    \end{tabular}%
  \label{tab:conf2}%
  \vspace{-1.5em}
\end{table}%

\subsection{QE-subnet}\label{cnn}

Given the compensated PQFs, the quality of non-PQFs can be enhanced through the QE-subnet, which is designed with spatio-temporal architecture. Specifically, together with the current processed non-PQF $F_{np}$, the compensated previous and subsequent PQFs (denoted by $F'_{p1}$ and $F'_{p2}$) are input to the QE-subnet. This way, both the spatial and temporal features of these three frames are explored and merged. Consequently, the advantageous information in the adjacent PQFs can be used to enhance the quality of the non-PQF. It differs from the CNN-based image/single-frame quality enhancement approaches, which only handle the spatial information within one frame.

\textbf{Architecture.} The architecture of the QE-subnet is shown in Figure \ref{qen}, and the details of the convolutional layers are presented in Table \ref{tab:conf2}. In the QE-subnet, the convolutional layers Conv 1, 2 and 3 are applied to extract the spatial features of input frames $F'_{p1}$, $F_{np}$ and $F'_{p2}$, respectively. Then, in order to use the high quality information of $F'_{p1}$, Conv 4 is adopted to merge the features of $F_{np}$ and $F'_{p1}$. That is, the outputs of Conv 1 and 2 are concatenated and then convolved by Conv 4. Similarly, Conv 5 is used to merge the features of $F_{np}$ and $F'_{p2}$. Conv 6/7 is designed to extract more complex features from Conv 4/5. Consequently, the extracted features of Conv 6 and Conv 7 are non-linearly mapped to another space through Conv 8. Finally, the reconstructed residual, denoted as $R_{np}(\theta_{qe})$, is achieved in Conv 9, and the non-PQF is enhanced by adding $R_{np}(\theta_{qe})$ to the input non-PQF $F_{np}$. Here, $\theta_{qe}$ is defined as the trainable parameters of QE-subnet.

\textbf{Training strategy.} The MC-subnet and QE-subnet of our MF-CNN are trained jointly in an end-to-end manner. Assume that $F'^R_{p1}$ and $F'^R_{p2}$ are defined as the raw frames of the previous and incoming PQFs, respectively. The loss function of our MF-CNN can be formulated as
\vspace{-.5em}
\begin{eqnarray}\label{lossmf}
L_{\text{MF}}(\theta_{mc},\theta_{qe}) = a\cdot\underbrace{\sum_{i=1}^2||F'^R_{pi}(\theta_{mc})-F^R_{np}||_2^2}_{L_{\text{MC}}:\ \text{loss of MC-subnet}} \nonumber \\
+ b\cdot\underbrace{\big|\big|\big(F_{np}+R_{np}(\theta_{qe})\big)-F^{R}_{np}\big|\big|_2^2}_{L_{\text{QE}}:\ \text{loss of QE-subnet}}.
\end{eqnarray}

As \eqref{lossmf} indicates, the loss function of the MF-CNN is the weighted sum of $L_{\text{MC}}$ and $L_{\text{QE}}$, which are the loss functions of MC-subnet and QE-subnet, respectively. Because $F'_{p1}$ and $F'_{p2}$ generated by the MC-subnet are the basis of the following QE-subnet, we set $a \gg b$ at the beginning of training. After the convergence of $L_{\text{MC}}$ is observed, we set $a\ll b$ to minimize the MSE between $F_{np}+R_{np}$ and $F^R_{np}$. As a result, the quality of non-PQF $F_{np}$ can be enhanced by using the high quality information of its nearest PQFs.

\vspace{-.5em}

\section{Experiments}\label{ex}

\subsection{Settings}\label{settings}

%
%\begin{table}[!t]
%  \centering
%  \footnotesize
%  \caption{Performance of the PQF detector on the test sequences.}
%    \begin{tabular}{|c|c|l|c|c|c|}
%    \hline
%      & \tabincell{c}{Our SVM-based PQF detector}&QP  &
%      \multicolumn{1}{c|}{Sequence}  &
%      Precision &
%      Recall &
%      $F_1$-score
%      \\
%    \hline
%      \multirow{11}{*}{37}&\textit{PeopleOnStreet} &
%      72.55\% &
%      97.37\%&
%      83.15\%
%      \\
%    \cline{2-5}
%&
%    \textit{TunnelFlag} &
%      100.0\%  &
%      98.68\% &
%      99.34\%
%      \\
%    \cline{2-5}
%
%    & \textit{Kimono} &
%      98.36\%  &
%      100.0\%  &
%      99.17\%
%      \\
%    \cline{2-5}
%     & \textit{BarScene} &
%      88.74\%  &
%      86.45\%  &
%      87.58\%
%      \\
%    \cline{2-5}
%    &
%    \textit{Vidyo1} &
%      98.68\%  &
%      100.0\%  &
%      99.34\%
%      \\
%    \cline{2-5}
%    &  \textit{Vidyo3}& 91.56\%& 94.00\%& 92.76\%\\
%    \cline{2-5}
%    &  \textit{Vidyo4}& 92.72\%& 74.47\%& 82.60\%\\
%    \cline{2-5}
%   &
%    \textit{RaceHorses} &
%      85.54\% &
%      92.21\%  &
%      88.75\%
%      \\
%    \cline{2-5}
%    & \textit{BasketballPass} &
%      93.08\%  &
%      92.37\%  &
%      92.72\%
%      \\
%    \cline{2-5}
%    &
%    \textit{MaD} &
%      85.53\% &
%      85.53\% &
%      85.53\%
%      \\
%    \cline{2-5}
%    & {\textbf{Average}} &
%      90.68\%&
%    92.11\%&
%    91.09\%
%     \\
%    \hline
%  \hline
%    42 &{\textbf{Average}} &
%    93.98\%  &
%      90.86\% &
%      92.23\%
%     \\
%    \hline
%    \end{tabular}%
%  \label{sc}%
%\end{table}%

The experimental results are presented to validate the effectiveness of our MFQE approach.
%, which is evaluated by several sequences compressed by the latest HEVC standard.
In our experiments, all 70 video sequences of our database introduced in Section \ref{quality} are randomly divided into the training set (60 sequences) and the test set (10 sequences). The training and test sequences are compressed by the latest HEVC standard, setting the Quantization Parameter (QP) to 42 and 37. We train two models of our MFQE approach for the sequences compressed at QP = 37 and 42, respectively. In the SVM-based PQF detector, the parameter $D$\footnote{$D$ should be changed according to coding standard and configurations.} is set to 6 in \eqref{c2}. It is because the maximal separation between two nearest PQFs is 6 frames in all training sequences compressed by HEVC. When training the MF-CNN, the raw and compressed sequences are segmented into $64\times64$ patches as the training samples. The batch size is set to 64. We adopt the Adam algorithm \cite{Kingma2014Adam} with initial learning rate as $10^{-4}$ to minimize the loss function of \eqref{lossmf}. In the training stage, we initially set $a = 1$ and $b = 0.01$ of \eqref{lossmf} to train the MC-subnet. After the MC-subnet converges, these hyperparameters are set as $a = 0.01$ and $b = 1$ to train the QE-subnet.

\subsection{Performance of the PQF detector}\label{ACC}
%
%Since PQF detection is the first stage in the proposed MFQE approach, the performance of our PQF detector is evaluated in terms of precision, recall and $F_1$-score, as shown in Table \ref{sc}. We can see from Table \ref{sc} that at QP = 37, the average precision and recall of our SVM-based PQF detector are $90.68\%$ and $92.11\%$ , respectively. In addition, the $F_1$-score, which is defined as the harmonic average of the precision and the recall, is $91.09\%$ on average. Similar results can be found at QP = 42, in which the average precision, recall and $F_1$-score are $93.98\%$, $90.86\%$ and $92.23\%$, respectively. Hence, the effectiveness of our SVM-based PQF detector is validated. Note that the No-Reference Image Quality Assessment (NR-IQA) approaches are not suitable for PQF detection, because they do not take the temporal information and prior knowledge of PQFs into consideration. For example, as shown in Table \ref{sc}, the latest NR-IQA approach \cite{bosse2017deep} can only achieve the $F_1$-score by $34.76\%$ and $29.77\%$ at QP = 37 and 42, respectively.
%
%Thus, we compare our results with the latest no-reference quality assessment method WaDIQaM-NR \cite{bosse2017deep}, in which the PQFs can be located using the estimated quality score of each frame.

Because PQF detection is the first stage of the proposed MFQE approach, the performance of our PQF detector is evaluated in terms of precision, recall and $F_1$-score, as shown in Table \ref{sc}. We can see from Table \ref{sc} that at QP = 37, the average precision and recall of our SVM-based PQF detector are $90.68\%$ and $92.11\%$ , respectively. In addition, the $F_1$-score, which is defined as the harmonic average of the precision and the recall, is $91.09\%$ on average. Similar results can be found for all test sequences compressed at QP = 42, in which the average precision, recall and $F_1$-score are $93.98\%$, $90.86\%$ and $92.23\%$, respectively. Thus, the effectiveness of our SVM-based PQF detector is validated.

\begin{table}[!t]
  \centering
  \footnotesize
  \caption{Performance of the PQF detector on the test sequences.}
    \begin{tabular}{|c|c|c|c|c|}
    \hline
     Method&
      QP &
      Precision &
      Recall &
      $F_1$-score
      \\
    \hline
    Our SVM-based &
      37 &
      \textbf{90.68\%}&
      \textbf{92.11\%} &
      \textbf{91.09\%}
      \\
\cline{2-5} PQF detector &
      42 &
      \textbf{93.98\%} &
      \textbf{90.86\%} &
      \textbf{92.23\%}
      \\
    \hline
  %  \multirow{2}[2]{*}{\cite{bosse2017deep}} &
%      37 &
%      33.92\% &
%      37.29\% &
%      35.37\%
%      \\
%\cline{2-5}     &
%      42 &
%      29.88\% &
%      31.10\% &
%      30.43\%
%      \\
%    \hline
    \end{tabular}%
  \label{sc}%
   \vspace{-1em}
\end{table}%

\begin{figure}[!t]
\includegraphics[width = 1\linewidth]{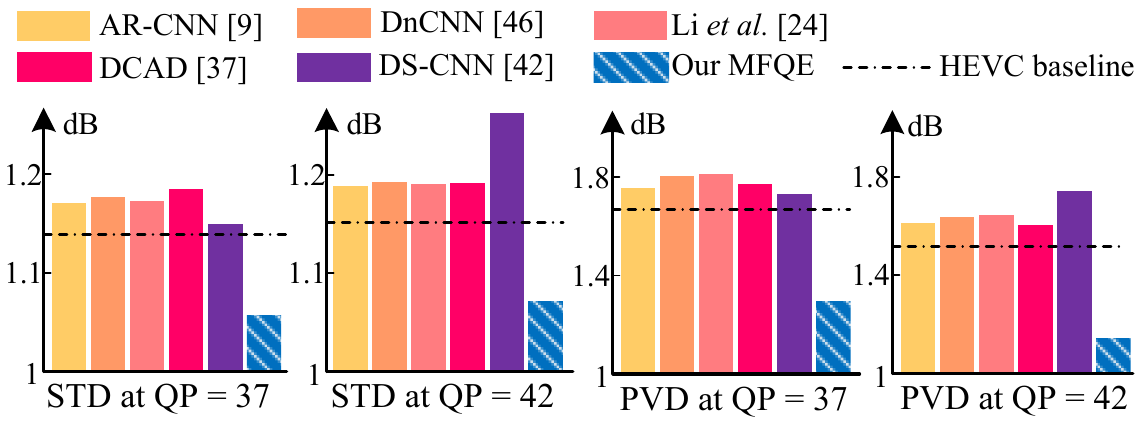}
\vspace{-1.5em}
\caption{\small{Averaged STD and PVD values of the test sequences.}}\label{stdpvd}
 \vspace{-.5em}
\end{figure}

\begin{figure}[!t]
\centering
\includegraphics[width = 1\linewidth]{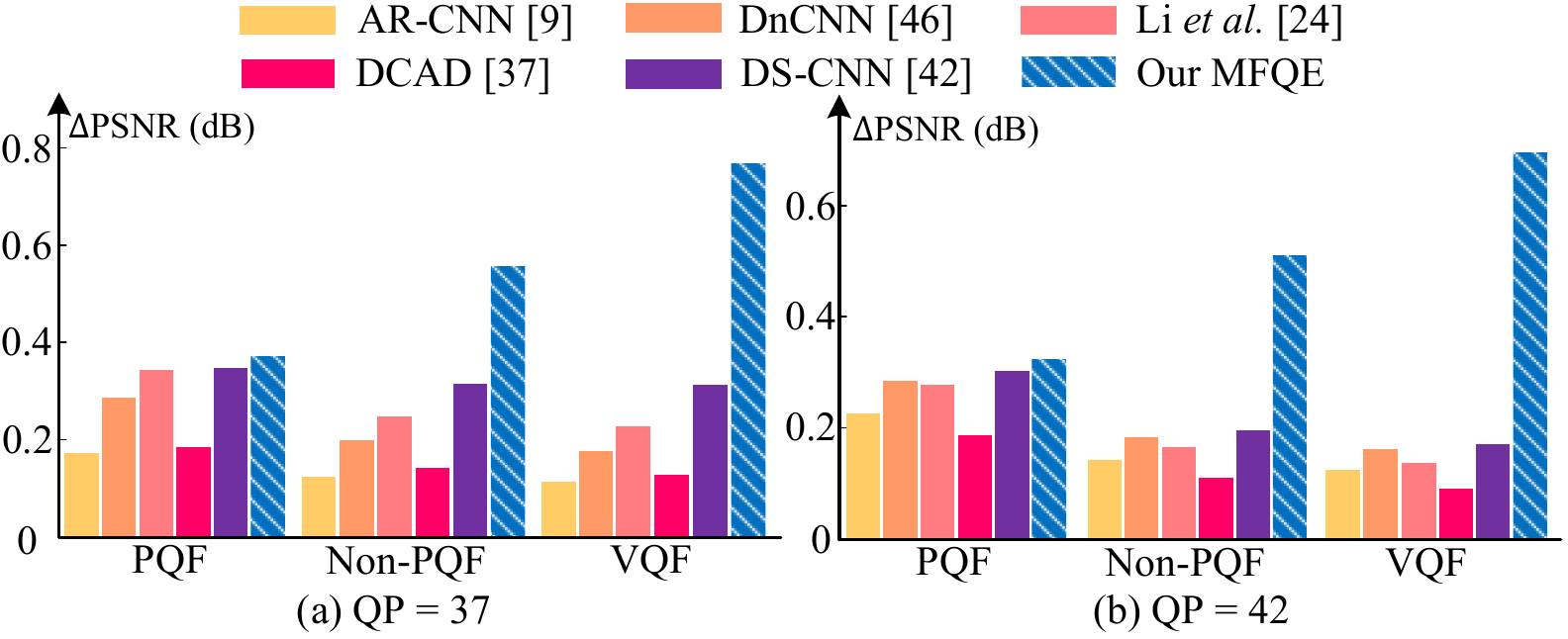}
\vspace{-1.5em}
\caption{\small{$\Delta$PSNR (dB) on PQFs, non-PQFs and VQFs of the test sequences.}}\label{npqf}
\vspace{-1.5em}
\end{figure}
%
%To our best knowledge, there is no other work on detecting PQFs. The nearest related works are No-Reference Image Quality Assessment (NR-IQA) methods, in which the PQFs can be located by the estimated quality score of each frame. However, since the NR-IQA methods do not consider any temporal information and prior knowledge of PQFs, they are not suitable for PQF detection. As Table \ref{sc} shows, the latest NR-IQA method \cite{bosse2017deep} performs much worse than our detector in classifying PQFs.
%
%
%We can further see that our PQF detector performs much better than \cite{bosse2017deep}, because the method of \cite{bosse2017deep} . In summary, the effectiveness of our SVM-based PQF detector is validated.

\subsection{Performance of our MFQE approach}\label{per}

In this section, we evaluate the quality enhancement performance of our MFQE approach in terms of $\Delta$PSNR, which measures the PSNR difference between the enhanced and the original compressed sequence. Our performance is compared with AR-CNN \cite{dong2015compression}, DnCNN \cite{Zhang2017Beyond}, Li \textit{et al.} \cite{li2017efficient}\footnote{Note that AR-CNN \cite{dong2015compression}, DnCNN \cite{Zhang2017Beyond} and Li \textit{et al.} \cite{li2017efficient} are re-trained on HEVC compressed samples for fair comparison.}, DCAD \cite{Wang2017A} and DS-CNN \cite{yang2017decoder}. Among them, AR-CNN, DnCNN and Li \textit{et al.} are the latest quality enhancement approaches for compressed image. DCAD and DS-CNN are the state-of-the-art video quality enhancement approaches.

\begin{table}
  \centering
  \scriptsize
  \caption{Overall $\Delta$PSNR (dB) of the test sequences.}
    \begin{tabular}{|>{\centering}p{.1cm}|>{\centering}p{.1cm}|>{\centering}p{.78cm}|>{\centering}p{.73cm}|>{\centering}p{.73cm}|>{\centering}p{.7cm}|>{\centering}p{.75cm}|p{.7cm}<{\centering}|}
    \hline
    \multicolumn{1}{|c|}{\hspace{-.5em} \multirow{1}[5]{*}{QP}}  &
    \multicolumn{1}{c|}{\hspace{-.5em} \multirow{1}[5]{*}{Seq.}} &
       \hspace{-.7em}{AR-CNN \cite{dong2015compression}} &
       \hspace{-.7em}{DnCNN \cite{Zhang2017Beyond}}&
       \hspace{-.7em}{Li \textit{et al.} \cite{li2017efficient}} &
       {DCAD \cite{Wang2017A}} &
      \hspace{-.7em}{DS-CNN} \cite{yang2017decoder}&
      {MFQE (our)}\\
    \hline
    \multicolumn{1}{|c|}{\hspace{-.5em} \multirow{10}[2]{*}{37}} &
      \multicolumn{1}{c|}{1} &

      0.1287 & 0.1955 &
      0.2523 &
      0.1354 &
      0.4762 &
      \textbf{0.7716}
      \\
\cline{2-8}    \multicolumn{1}{|c|}{} &
      \multicolumn{1}{c|}{2} &
      0.0718 & 0.1888 &
      0.2857 &
      0.0376 &
      0.4228 &
      \textbf{0.6042}
      \\
\cline{2-8}    \multicolumn{1}{|c|}{} &
      \multicolumn{1}{c|}{3} &
      0.1095 & 0.1328 &
      0.1872 &
      0.1112 &
      0.2394 &
      \textbf{0.4715}
      \\
\cline{2-8}    \multicolumn{1}{|c|}{} &
      \multicolumn{1}{c|}{4} &
      0.1304 & 0.2084 &
      0.2170 &
      0.0796 &
      0.3173 &
      \textbf{0.4381}
      \\
\cline{2-8}    \multicolumn{1}{|c|}{} &
      \multicolumn{1}{c|}{5} &
      0.1900 & 0.2936 &
      0.3645 &
      0.2334 &
      0.3252 &
      \textbf{0.5496}
      \\
\cline{2-8}    \multicolumn{1}{|c|}{} &
      \multicolumn{1}{c|}{6} &
      0.1522 & 0.1944 &
      0.2630 &
      0.1619 &
      0.3728 &
      \textbf{0.5980}
      \\
\cline{2-8}    \multicolumn{1}{|c|}{} &
      \multicolumn{1}{c|}{7} &
      0.1445 & 0.2224 &
      0.2570 &
      0.1775 &
      0.2777 &
      \textbf{0.3898}
      \\
\cline{2-8}    \multicolumn{1}{|c|}{} &
      \multicolumn{1}{c|}{8} &
      0.1305 & 0.2424 &
      0.2939 &
      0.1940 &
      0.2790 &
      \textbf{0.4838}
      \\
\cline{2-8}    \multicolumn{1}{|c|}{} &
      \multicolumn{1}{c|}{9} &
      0.1573 & 0.2588 &
      0.3034 &
      0.2224 &
      0.2720 &
      \textbf{0.3935}
      \\
\cline{2-8}    \multicolumn{1}{|c|}{} &
      \multicolumn{1}{c|}{10} &
      0.1490 & 0.2509 &
      0.2926 &
      0.2026 &
      0.2498 &
      \textbf{0.4019}
       \\
\cline{2-8}    \multicolumn{1}{|c|}{} &
      \multicolumn{1}{c|}{\textbf{Ave.}} &
      0.1364 & 0.2188 &
      0.2717 &
      0.1556 &
      0.3232 &
      \textbf{0.5102}
       \\
    \hline
    \hline
    \multicolumn{1}{|c|}{\hspace{-.5em} \multirow{1}[2]{*}{42}} &\textbf{Ave.}&0.1627&0.2073&0.1924&0.1282&0.2189&\textbf{0.4610}
       \\
    \hline
    \multicolumn{8}{c}{1: \textit{PeopleOnStreet}\ \ 2: \textit{TunnelFlag}\ \ 3: \textit{Kimono}\ \ 4: \textit{BarScene}\ \ 5: \textit{Vidyo1}}\\
    \multicolumn{8}{c}{6: \textit{Vidyo3}\ \ 7: \textit{Vidyo4}\ \ 8: \textit{BasketballPass}\ \ 9: \textit{RaceHorses}\ \ 10: \textit{MaD}}
    \end{tabular}%
  \label{tab:dpsnr}%
  \vspace{-1.5em}
\end{table}%

\textbf{Quality enhancement on non-PQFs.} Our MFQE approach mainly focuses on enhancing the quality of non-PQFs using the multi-frame information. Therefore, we first assess the quality enhancement of non-PQFs. Figure \ref{npqf} shows the $\Delta$PSNR results averaged over PQFs, non-PQFs and VQFs of all 10 test sequences, compressed at QP = 37. As shown in this figure, our MFQE approach has a considerably larger PSNR improvement for non-PQFs, compared to that for PQFs. Furthermore, an even higher $\Delta$PSNR can be achieved for VQFs in our approach. In contrast, for compared approaches, the PSNR improvement of non-PQFs is similar to or even less than that of PQFs. Specifically, for non-PQFs, our MFQE approach doubles $\Delta$PSNR of DS-CNN \cite{yang2017decoder}, which performs best among all of the compared approaches. This validates the effectiveness of our MFQE approach in enhancing the quality of non-PQFs.

\begin{figure}[!t]
\subfigure{\includegraphics[width = 1\linewidth]{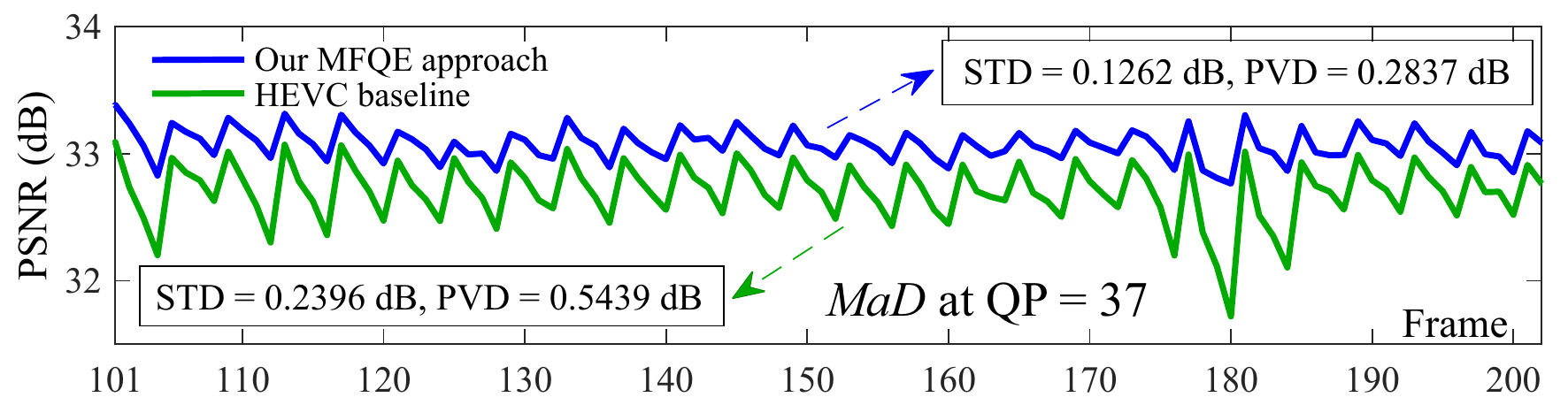}} \nonumber
\subfigure{\includegraphics[width = 1\linewidth]{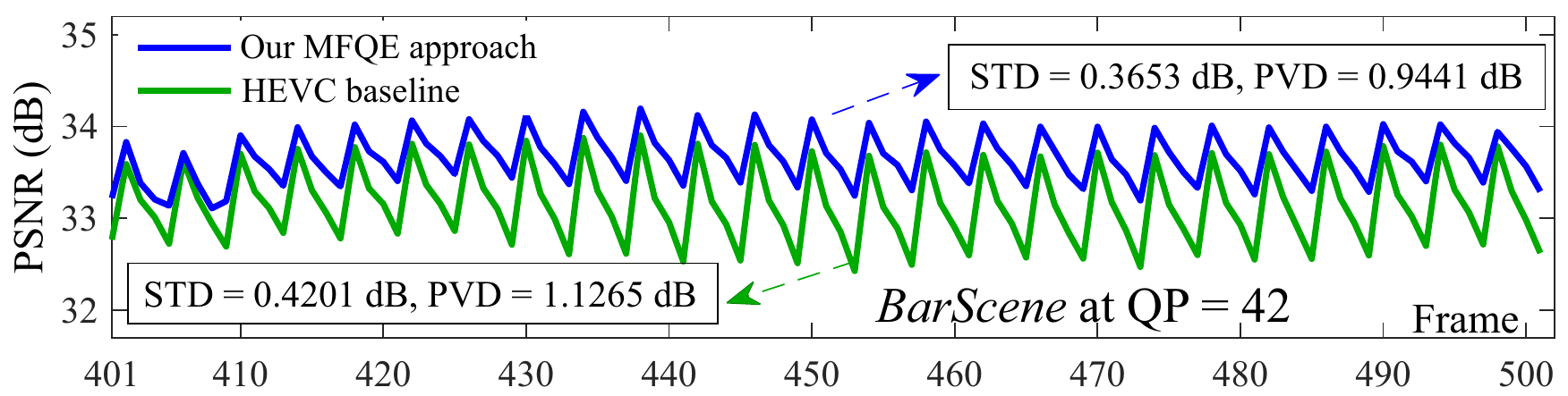}}\nonumber
\caption{\small{PSNR curves of HEVC and our MFQE approach.}}\label{curve}
 \vspace{-1.5em}
\end{figure}

\begin{figure*}[!t]
\centering
\vspace{-1.5em}
\subfigure{\includegraphics[width = 1\linewidth]{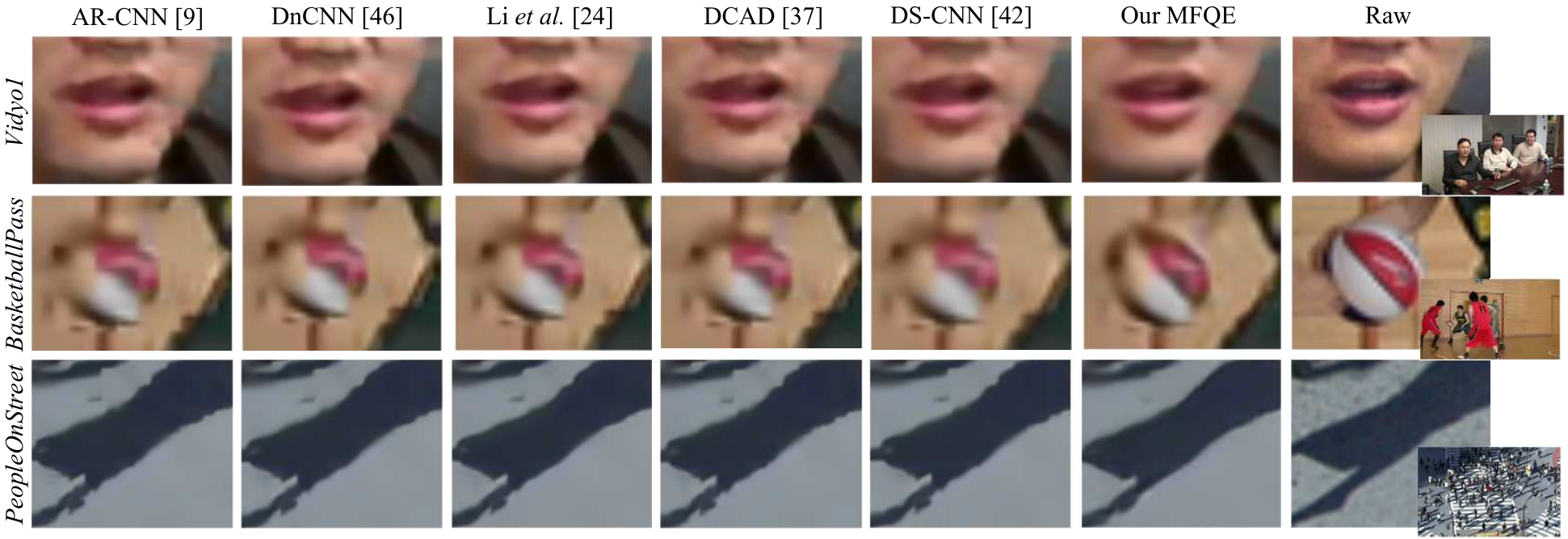}}
\vspace{-1.5em}
\caption{\small{Subjective quality performance on \textit{Vidyo1} at QP = 37, \textit{BasketballPass} at QP = 37 and \textit{PeopleOnStreet} at QP = 42.}}\label{sq}
\vspace{-1.5em}
\end{figure*}

\textbf{Overall quality enhancement.} Table \ref{tab:dpsnr} presents the $\Delta$PSNR results averaged over all frames, for each test sequence. As this table shows, our MFQE approach outperforms all five compared approaches for all test sequences. To be specific, at QP = 37, the highest $\Delta$PSNR of our MFQE approach reaches 0.7716 dB. The averaged $\Delta$PSNR of our MFQE approach is 0.5102 dB, which is $87.78\%$ higher than that of of Li \textit{et al.} \cite{li2017efficient} (0.2717 dB), and $57.86\%$ higher than that of DS-CNN (0.3233 dB). Besides, more $\Delta$PSNR gain can be obtained in our MFQE approach, when compared with AR-CNN \cite{dong2015compression}, DnCNN \cite{Zhang2017Beyond} and DCAD \cite{Wang2017A}. At QP = 42, our MFQE approach ($\Delta$PSNR = 0.4610 dB) also doubles the PSNR improvement of the second best approach DS-CNN ($\Delta$PSNR = 0.2189 dB). Thus, our MFQE approach is effective in overall quality enhancement. This is mainly due to the large improvement in non-PQFs, which are the majority of compressed video frames.

\textbf{Quality fluctuation.} Apart from the artifacts, the quality fluctuation of compressed video may also lead to degradation of QoE \cite{He2001Low,Vito2005PSNR,Hu2012Adaptive}. Fortunately, as discussed above, our MFQE approach is able to mitigate the quality fluctuation, because of the higher PSNR improvement of non-PQFs. We evaluate the fluctuation of video quality in terms of the STD and PVD of the PSNR curve, as introduced in Section \ref{quality}. Figure \ref{stdpvd} shows the STD and PVD values averaged over all test sequences, for the HEVC baseline and the quality enhancement approaches. As shown in this figure, our MFQE approach succeeds in reducing the STD and PVD after enhancing the quality of the compressed sequences. By contrast, the five compared approaches enlarge the STD and PVD values over the HEVC baseline. Thus, our MFQE approach is able to mitigate the quality fluctuation and achieve better QoE, compared with other approaches. Figure \ref{curve} further shows the PSNR curves of the HEVC baseline and our MFQE approach for two test sequences. It can be seen that the PSNR fluctuation of our MFQE approach is obviously less than the HEVC baseline. To summarize, our MFQE approach is effective to mitigate the quality fluctuation of compressed video, meanwhile enhancing video quality.

\textbf{Subjective quality performance.} Figure \ref{sq} shows the subjective quality performance on the sequences \textit{Vidyo1} at QP = 37, \textit{BasketballPass} at QP = 37 and \textit{PeopleOnStreet} at QP = 42. One may observe from Figure \ref{sq} that our MFQE approach reduces the compression artifacts more effectively than the five compared approaches. Specifically, the severely distorted content, e.g., the mouth in \textit{Vidyo1}, the ball in \textit{BasketballPass} and the shadow in \textit{PeopleOnStreet}, can be finely restored in our MFQE approach upon the same content from the neighboring high quality frames. In contrast, such distortion can hardly be restored in the compared approaches, which only use the single low quality frame.

\textbf{Effectiveness of utilizing PQFs.} Finally, we validate the effectiveness of utilizing PQFs by re-training our MF-CNN to enhance non-PQFs using adjacent frames instead of PQFs. In our experiments, using adjacent frames instead of PQFs only has 0.3896 dB and 0.3128 dB $\Delta$PSNR at QP=37 and 42, respectively. By contrast, as aforementioned, utilizing PQFs achieves $\Delta$PSNR = 0.5102 dB and 0.4610 dB at QP = 37 and 42. Moreover, it has been discussed before that non-PQFs, which are enhanced taking advantage of PQFs, have much larger $\Delta$PSNR than PQFs. These validate the effectiveness of utilizing PQFs in our MFQE approach.

\subsection{Transfer to H.264 standard}\label{trans}

We further verify the generalization capability of our MFQE approach by transferring to H.264 compressed sequences. The training and test sequences of our database are compressed by H.264 at QP = 37. Then, the training sequences are used to fine-tune the MF-CNN model. Then, the quality of the test H.264 sequences is improved by our MFQE approach with the fine-tuned MF-CNN model. We find that the average PSNR of test sequences can be increased by 0.4540 dB. This result is comparable to that of HEVC (0.5102 dB). Therefore, the generalization capability of our MFQE approach can be verified.
\vspace{-.5em}
\section{Conclusion}

In this paper, we have proposed a CNN-based MFQE approach to reduce compression artifacts of video. Differing from the conventional single frame quality enhancement, our MFQE approach improves the quality of one frame by using high quality content of its nearest PQFs. To this end, we developed an SVM-based PQF detector to classify PQFs and non-PQFs in compressed video. Then, we proposed a novel CNN framework, called MF-CNN, to enhance the quality of each non-PQF. Specifically, the MC-subnet of our MF-CNN compensates motion between PQFs and non-PQFs. Subsequently, the QE-subnet enhances the quality of each non-PQF by inputting the current non-PQF and the nearest compensated PQFs. Finally, experimental results showed that our MFQE approach significantly improves the quality of non-PQFs, far better than other state-of-the-art quality enhancement approaches. Consequently, the overall quality enhancement is considerably higher than other approaches, with less quality fluctuation.
\vspace{-.5em}
\section*{Acknowledgement}
This work was supported by the NSFC projects under Grants 61573037 and 61202139, and Fok Ying-Tong education foundation under Grant 151061.

{\small
\bibliographystyle{ieee}
\bibliography{egbib}

\begin{thebibliography}{10}\itemsep=-1pt

\bibitem{bossen2011common}
F.~Bossen.
\newblock Common test conditions and software reference configurations.
\newblock In {\em Joint Collaborative Team on Video Coding (JCT-VC) of ITU-T
  SG16 WP3 and ISO/IEC JTC1/SC29/WG11, 5th meeting, Jan. 2011}, 2011.

\bibitem{brandi2008super}
F.~Brandi, R.~de~Queiroz, and D.~Mukherjee.
\newblock Super resolution of video using key frames.
\newblock In {\em Proceedings of the IEEE International Symposium on Circuits
  and Systems (ISCAS)}, pages 1608--1611. IEEE, 2008.

\bibitem{Caballero_2017_CVPR}
J.~Caballero, C.~Ledig, A.~Aitken, A.~Acosta, J.~Totz, Z.~Wang, and W.~Shi.
\newblock Real-time video super-resolution with spatio-temporal networks and
  motion compensation.
\newblock In {\em Proceedings of the IEEE Conference on Computer Vision and
  Pattern Recognition (CVPR)}, July 2017.

\bibitem{Cavigelli2017CAS}
L.~Cavigelli, P.~Hager, and L.~Benini.
\newblock {CAS-CNN}: A deep convolutional neural network for image compression
  artifact suppression.
\newblock In {\em Proceedings of the International Joint Conference on Neural
  Networks (IJCNN)}, pages 752--759, 2017.

\bibitem{CC01a}
C.-C. Chang and C.-J. Lin.
\newblock {LIBSVM}: A library for support vector machines.
\newblock {\em ACM Transactions on Intelligent Systems and Technology},
  2:27:1--27:27, 2011.

\bibitem{chang2014reducing}
H.~Chang, M.~K. Ng, and T.~Zeng.
\newblock Reducing artifacts in {JPEG} decompression via a learned dictionary.
\newblock {\em IEEE Transactions on Signal Processing}, 62(3):718--728, 2014.

\bibitem{Cisco}
I.~Cisco~Systems.
\newblock Cisco visual networking index: Global mobile data traffic forecast
  update.
\newblock
  https://www.cisco.com/c/en/us/solutions/collateral/service-provider/visual-networking-index-vni/mobile-white-paper-c11-520862.html.

\bibitem{dai2017convolutional}
Y.~Dai, D.~Liu, and F.~Wu.
\newblock A convolutional neural network approach for post-processing in {hevc}
  intra coding.
\newblock In {\em Proceedings of the International Conference on Multimedia
  Modeling (MMM)}, pages 28--39. Springer, 2017.

\bibitem{dong2015compression}
C.~Dong, Y.~Deng, C.~Change~Loy, and X.~Tang.
\newblock Compression artifacts reduction by a deep convolutional network.
\newblock In {\em Proceedings of the IEEE International Conference on Computer
  Vision (ICCV)}, pages 576--584, 2015.

\bibitem{foi2007pointwise}
A.~Foi, V.~Katkovnik, and K.~Egiazarian.
\newblock Pointwise shape-adaptive {DCT} for high-quality denoising and
  deblocking of grayscale and color images.
\newblock {\em IEEE Transactions on Image Processing}, 16(5):1395--1411, 2007.

\bibitem{Gall1992The}
D.~J.~L. Gall.
\newblock The {MPEG} video compression algorithm.
\newblock {\em Signal Processing: Image Communication}, 4(2):129¨C140, 1992.

\bibitem{Guo2016Building}
J.~Guo and H.~Chao.
\newblock Building dual-domain representations for compression artifacts
  reduction.
\newblock In {\em Proceedings of the European Conference on Computer Vision
  (ECCV)}, pages 628--644, 2016.

\bibitem{he2016deep}
K.~He, X.~Zhang, S.~Ren, and J.~Sun.
\newblock Deep residual learning for image recognition.
\newblock In {\em Proceedings of the IEEE conference on Computer Vision and
  Pattern Recognition (CVPR)}, pages 770--778, 2016.

\bibitem{He2016Delving}
K.~He, X.~Zhang, S.~Ren, and J.~Sun.
\newblock Delving deep into rectifiers: Surpassing human-level performance on
  imagenet classification.
\newblock In {\em Proceedings of the IEEE International Conference on Computer
  Vision (ICCV)}, pages 1026--1034, 2016.

\bibitem{He2001Low}
Z.~He, Y.~K. Kim, and S.~K. Mitra.
\newblock Low-delay rate control for {DCT} video coding via $\rho$-domain
  source modeling.
\newblock {\em IEEE Transactions on Circuits and Systems for Video Technology},
  11(8):928--940, 2001.

\bibitem{Hu2012Adaptive}
S.~Hu, H.~Wang, and S.~Kwong.
\newblock Adaptive quantization-parameter clip scheme for smooth quality in
  {H}.264/{AVC}.
\newblock {\em IEEE Transactions on Image Processing}, 21(4):1911--1919, 2012.

\bibitem{Huang2015Bidirectional}
Y.~Huang, W.~Wang, and L.~Wang.
\newblock Bidirectional recurrent convolutional networks for multi-frame
  super-resolution.
\newblock In {\em Proceedings of the Advances in Neural Information Processing
  Systems (NIPS)}, pages 235--243, 2015.

\bibitem{jancsary2012loss}
J.~Jancsary, S.~Nowozin, and C.~Rother.
\newblock Loss-specific training of non-parametric image restoration models: A
  new state of the art.
\newblock In {\em Proceedings of the European Conference on Computer Vision
  (ECCV)}, pages 112--125. Springer, 2012.

\bibitem{jung2012image}
C.~Jung, L.~Jiao, H.~Qi, and T.~Sun.
\newblock Image deblocking via sparse representation.
\newblock {\em Image Communication}, 27(6):663--677, 2012.

\bibitem{Kappeler2016Video}
A.~Kappeler, S.~Yoo, Q.~Dai, and A.~K. Katsaggelos.
\newblock Video super-resolution with convolutional neural networks.
\newblock {\em IEEE Transactions on Computational Imaging}, 2(2):109--122,
  2016.

\bibitem{Kingma2014Adam}
D.~P. Kingma and J.~Ba.
\newblock Adam: A method for stochastic optimization.
\newblock {\em Computer Science}, 2014.

\bibitem{lecun1998gradient}
Y.~LeCun, L.~Bottou, Y.~Bengio, and P.~Haffner.
\newblock Gradient-based learning applied to document recognition.
\newblock {\em Proceedings of the IEEE}, 86(11):2278--2324, 1998.

\bibitem{Li2017Video}
D.~Li and Z.~Wang.
\newblock Video super-resolution via motion compensation and deep residual
  learning.
\newblock {\em IEEE Transactions on Computational Imaging}, PP(99):1--1, 2017.

\bibitem{li2017efficient}
K.~Li, B.~Bare, and B.~Yan.
\newblock An efficient deep convolutional neural networks model for compressed
  image deblocking.
\newblock In {\em Proceedings of the IEEE International Conference on
  Multimedia and Expo (ICME)}, pages 1320--1325. IEEE, 2017.

\bibitem{li2015novel}
S.~Li, M.~Xu, and Z.~Wang.
\newblock A novel method on optimal bit allocation at {LCU} level for rate
  control in {HEVC}.
\newblock In {\em Proceedings of the IEEE International Conference on
  Multimedia and Expo (ICME)}, pages 1--6. IEEE, 2015.

\bibitem{li2017optimal}
S.~Li, M.~Xu, Z.~Wang, and X.~Sun.
\newblock Optimal bit allocation for {CTU} level rate control in {HEVC}.
\newblock {\em IEEE Transactions on Circuits and Systems for Video Technology},
  27(11):2409--2424, 2017.

\bibitem{liew2004blocking}
A.-C. Liew and H.~Yan.
\newblock Blocking artifacts suppression in block-coded images using
  overcomplete wavelet representation.
\newblock {\em IEEE Transactions on Circuits and Systems for Video Technology},
  14(4):450--461, 2004.

\bibitem{Makansi2017End}
O.~Makansi, E.~Ilg, and T.~Brox.
\newblock End-to-end learning of video super-resolution with motion
  compensation.
\newblock In {\em Proceedings of the German Conference on Pattern Recognition
  (GCPR)}, pages 203--214, 2017.

\bibitem{Mittal2012No}
A.~Mittal, A.~K. Moorthy, and A.~C. Bovik.
\newblock No-reference image quality assessment in the spatial domain.
\newblock {\em IEEE Transactions on Image Processing}, 21(12):4695--4708, 2012.

\bibitem{Schafer1995Digital}
R.~Schafer and T.~Sikora.
\newblock Digital video coding standards and their role in video
  communications.
\newblock {\em Proceedings of the IEEE}, 83(6):907--924, 1995.

\bibitem{Sikora2002The}
T.~Sikora.
\newblock The {MPEG}-4 video standard verification model.
\newblock {\em IEEE Transactions on Circuits and Systems for Video Technology},
  7(1):19--31, 2002.

\bibitem{song2011video}
B.~C. Song, S.-C. Jeong, and Y.~Choi.
\newblock Video super-resolution algorithm using bi-directional overlapped
  block motion compensation and on-the-fly dictionary training.
\newblock {\em IEEE Transactions on Circuits and Systems for Video Technology},
  21(3):274--285, 2011.

\bibitem{sullivan2012overview}
G.~J. Sullivan, J.~Ohm, W.-J. Han, and T.~Wiegand.
\newblock Overview of the high efficiency video coding ({HEVC}) standard.
\newblock {\em IEEE Transactions on Circuits and Systems for Video Technology},
  22(12):1649--1668, 2012.

\bibitem{Vito2005PSNR}
F.~D. Vito and J.~C.~D. Martin.
\newblock Psnr control for {GOP}-level constant quality in {H}.264 video
  coding.
\newblock In {\em Proceedings of the IEEE International Symposium on Signal
  Processing and Information Technology}, pages 612--617, 2005.

\bibitem{vlachos2000detection}
T.~Vlachos.
\newblock Detection of blocking artifacts in compressed video.
\newblock {\em Electronics Letters}, 36(13):1106--1108, 2000.

\bibitem{wang2013adaptive}
C.~Wang, J.~Zhou, and S.~Liu.
\newblock Adaptive non-local means filter for image deblocking.
\newblock {\em Signal Processing: Image Communication}, 28(5):522--530, 2013.

\bibitem{Wang2017A}
T.~Wang, M.~Chen, and H.~Chao.
\newblock A novel deep learning-based method of improving coding efficiency
  from the decoder-end for {HEVC}.
\newblock In {\em Proceedings of the Data Compression Conference (DCC)}, 2017.

\bibitem{wang2016d3}
Z.~Wang, D.~Liu, S.~Chang, Q.~Ling, Y.~Yang, and T.~S. Huang.
\newblock {D}3: Deep dual-domain based fast restoration of {JPEG}-compressed
  images.
\newblock In {\em Proceedings of the IEEE Conference on Computer Vision and
  Pattern Recognition (CVPR)}, pages 2764--2772, 2016.

\bibitem{wiegand2003overview}
T.~Wiegand, G.~J. Sullivan, G.~Bjontegaard, and A.~Luthra.
\newblock Overview of the {H. 264/AVC} video coding standard.
\newblock {\em IEEE Transactions on Circuits and Systems for Video Technology},
  13(7):560--576, 2003.

\bibitem{Xiph}
Xiph.org.
\newblock Xiph.org video test media.
\newblock https://media.xiph.org/video/derf/.

\bibitem{yang2016subjective}
R.~Yang, M.~Xu, L.~Jiang, and Z.~Wang.
\newblock Subjective-quality-optimized complexity control for {HEVC} decoding.
\newblock In {\em Multimedia and Expo (ICME), 2016 IEEE International
  Conference on}, pages 1--6. IEEE, 2016.

\bibitem{yang2017decoder}
R.~Yang, M.~Xu, and Z.~Wang.
\newblock Decoder-side {HEVC} quality enhancement with scalable convolutional
  neural network.
\newblock In {\em Multimedia and Expo (ICME), 2017 IEEE International
  Conference on}, pages 817--822. IEEE, 2017.

\bibitem{yang2018saliency}
R.~Yang, M.~Xu, Z.~Wang, Y.~Duan, and X.~Tao.
\newblock Saliency-guided complexity control for {HEVC} decoding.
\newblock {\em IEEE Transactions on Broadcasting}, PP(99):1--18, 2018.

\bibitem{yang2017enhancing}
R.~Yang, M.~Xu, Z.~Wang, and Z.~Guan.
\newblock Enhancing quality for {HEVC} compressed videos.
\newblock {\em arXiv preprint arXiv:1709.06734}, 2017.

\bibitem{zeng2014characterizing}
K.~Zeng, T.~Zhao, A.~Rehman, and Z.~Wang.
\newblock Characterizing perceptual artifacts in compressed video streams.
\newblock In {\em Human Vision and Electronic Imaging XIX}. International
  Society for Optics and Photonics, 2014.

\bibitem{Zhang2017Beyond}
K.~Zhang, W.~Zuo, Y.~Chen, D.~Meng, and L.~Zhang.
\newblock Beyond a gaussian denoiser: Residual learning of deep cnn for image
  denoising.
\newblock {\em IEEE Transactions on Image Processing}, 26(7):3142--3155, 2017.

\end{thebibliography}
}

\end{document}